\documentclass[10pt,twocolumn,letterpaper]{article}

\usepackage{cvpr}
\usepackage{times}
\usepackage{epsfig}
\usepackage{graphicx}
\usepackage{amsmath}
\usepackage{amssymb}
\usepackage{enumitem}
\usepackage{float}

\usepackage[pagebackref=true,breaklinks=true,letterpaper=true,colorlinks,bookmarks=false]{hyperref}

\cvprfinalcopy 

\def\cvprPaperID{****} 

\ifcvprfinal\fi
\begin{document}

\title{HP-GAN: Probabilistic 3D human motion prediction via GAN}

\author{Emad Barsoum \qquad John Kender\\
Columbia University\\
116th St \& Broadway, New York, NY 10027\\
{\tt\small eb2871@columbia.edu} \qquad {\tt\small jrk@cs.columbia.edu}
\and
Zicheng Liu\\
Microsoft\\
One Microsoft Way, WA 98052\\
{\tt\small zliu@microsoft.com}
}

\maketitle


\begin{abstract}

Predicting and understanding human motion dynamics has many applications, such as motion synthesis, augmented reality, security, and autonomous vehicles. Due to the recent success of generative adversarial networks (GAN), there has been much interest in probabilistic estimation and synthetic data generation using deep neural network architectures and learning algorithms.

We propose a novel sequence-to-sequence model for probabilistic human motion prediction, trained with a modified version of improved Wasserstein generative adversarial networks (WGAN-GP), in which we use a custom loss function designed for human motion prediction. Our
model, which we call HP-GAN, learns a probability density function of future human poses conditioned on previous poses. It predicts multiple sequences of possible future human poses, each from the same input sequence but a different vector $z$ drawn from a random distribution. Furthermore, to quantify the quality of the non-deterministic predictions, we simultaneously train a motion-quality-assessment model that learns the probability that a given skeleton sequence is a real human motion.

We test our algorithm on two of the largest skeleton datasets: NTURGB-D and Human3.6M. We train our model on both single and multiple action types. Its predictive power for long-term motion estimation is demonstrated by generating multiple plausible futures of more than 30 frames from just 10 frames of input. We show that most sequences generated from the same input have more than 50\% probabilities of being judged as a real human sequence. We will release all the code used in this paper to Github.

\end{abstract}

\section{Introduction}

Accurate short-term (several second) predictions of what will happen in the world given past events is a fundamental and useful human ability. Such aptitude is vital for daily activities, social interactions, and ultimately survival. For example, driving requires predicting other cars' and pedestrians' motions in order to avoid an accident; handshaking requires predicting the location of the other person's hand; and playing sports requires predicting other players' reactions. In order to create a machine that can interact seamlessly with the world, it needs a similar ability to understand the dynamics of the human world, and to predict probable futures based on learned history and the immediate present.

However, the future is not deterministic, so predicting the future cannot be deterministic, except in the very short term. As the predictions extend further into the future, uncertainty becomes higher. People walking may turn or fall; people throwing a ball may drop it instead.  However, some predictions are more plausible than others, and have a higher probability.

In this paper, we focus on creating a model that can predict multiple plausible future human (skeleton) poses from a given past. The number of poses taken from the immediate past, and the predicted number of poses in the future, which can be unrestricted, are parameters to the model. To accomplish this, we use a modified version of the improved Wasserstein generative adversarial network (WGAN-GP)\cite{corr2017:Ishaan} with a custom loss function that takes into consideration human motion and human anatomy.

The generator is a novel adaptation of sequence-to-sequence model\cite{nips2014:Ilya} of poses derived from a Recurrent Neural Network (RNN), and the critic and discriminator are a multilayer network (MLP). We use the critic network to train the generator and the discriminator network to learn distinguishing between a real sequence of poses from a fake one. In essence, we combine some aspect of the original GAN~\cite{nips2014:Goodfellow} with WGAN-GP\cite{corr2017:Ishaan}. We train our model on all actions at once, so its output is not conditioned on any specific human action. Our model takes as input a sequence of previous skeletal poses, plus a random vector $z$ from the reduced sequence space which samples possible future poses. For each such $z$ value, the model generates a different output sequence of possible future poses.

We use an RNN for the generator because RNNs are a class of neural networks designed to model sequences, especially variable length sequences. They have been successfully used in machine translation \cite{nips2014:Ilya}, caption generation from images \cite{iccv2015:Jia, cvpr2015:Donahue}, video classification and action
recognition\cite{hbu2011:Baccouche, cvpr2015:Yue-Hei, cvpr2015:Donahue}, action detection\cite{cvpr2016:Yeung}, video description\cite{cvpr2016:Pan, corr2014:Kiros, corr2015:Yao, cvpr2015:Donahue}, sequence prediction\cite{corr2013:Graves, icml2015:Srivastava} and others.

We structure the learning by using a GAN because GANs\cite{nips2014:Goodfellow} are a class of unsupervised learning algorithms, inspired by game theory\cite{pjm1958:Sion}, which allow the generation of futures that are not tied to specific ground truth. Among other domains, they have been used to generate impressive realistic images.  
However, GANs have weaknesses. They can be difficult to train and unstable in their learning, their loss value does not necessarily indicate the quality of the generated sample, and the training can collapse easily. Recent literature\cite{corr2017:Arjovsky,corr2017:Ishaan,corr2017:Neyshabur,corr2016:Uehara,corr2017:Qi,corr2016:Mao} tries to improve GAN training and provide a theoretical guaranty for its convergence. In our work, we address this by adding a custom loss based on the skeleton physics in addition to the GAN loss, in order to stabilize and improve the training.

To quantitatively assess the quality of the non-deterministic predictions, we simultaneously train a motion-quality-assessment model that learns the probability that a given skeleton sequence is a real human motion.

We test our motion prediction model on two large datasets each captured with a different modality. The first is the NTURGB-D\cite{cvpr2016:Shahroudy} dataset, which is the largest available RGB-D and skeleton-based dataset, with data captured by using the Microsoft Kinect v2 sensor. The second is the Human3.6M\cite{pami2014:Ionescu,iccv2011:Ionescu} dataset, which is one of the largest available datasets derived from motion capture (MoCap) data.

Our main contributions are:
\setlist{nolistsep}
\begin{enumerate}[noitemsep]
\item We propose a novel human motion model that can predict multiple possible futures from a single past.
\item We propose a motion-quality-assessment model to quantitatively evaluate the quality of the predicted human motions.
\end{enumerate}

\section{Related work on prediction}

Since the introduction of the Kinect sensor, there has been much work on recognizing human action and predicting human poses from skeleton data. For example, predicting human poses conditioned on previous poses using deep RNNs~\cite{iccv2015:Katerina, cvpr2016:Ashesh, cvpr2017:julieta} is due in part to this availability of large human motion datasets~\cite{pami2014:Ionescu,iccv2011:Ionescu,cvpr2016:Shahroudy}. In general, human motion prediction can be categorized into two categories: probabilistic and deterministic prediction.

\subsection{Probabilistic motion prediction}

Most work in probabilistic human motion prediction uses non-deep learning approaches~\cite{nips2000:Pavlovic, eccv2002:Sidenbladh, nips2005:Wang, pami2008:Wang, icml2013:Hema, cvpr2014:Lehrmann, iccv2015:Katerina}. In~\cite{cvpr2014:Lehrmann}, the authors propose simple Markov models that model observed data, and use the proposed model for action recognition and task completion. The limitation in this approach is that it predicts motion from a single action only, and assumes that human motion satisfies the Markov assumption. In~\cite{pami2008:Wang}, the authors introduce Gaussian process dynamical models (GPDMs) to model human pose and motion. However, they train their model on each action separately, and constrain the model to a Gaussian process. In~\cite{eccv2002:Sidenbladh}, the authors map human motion to a low dimensional space, and use the coordinates as an index into a binary tree that predicts the next pose, thus casting the prediction problem into a search problem. However, this approach can not generalize to previously unseen motions. In~\cite{nips2000:Pavlovic}, the authors use switching linear dynamic systems learned through a Bayesian network, and in~\cite{icml2013:Hema} the author used conditional random fields (CRF) to model spatio-temporal dynamics.

In contrast to all the above, our work does not use any statistical models to constrain the motion. As far as we know, we are the first to use deep neural networks for probabilistic motion prediction.


\subsection {Deterministic motion prediction}

Recent human motion prediction, which relies on deep RNNs~\cite{iccv2015:Katerina, cvpr2016:Ashesh, cvpr2017:julieta} or deep neutral networks~\cite{corr2017:Judith}, is primarily deterministic. In~\cite{iccv2015:Katerina}, the authors mix both deterministic and probabilistic human motion predictions. Their deterministic aspect is based on a modified RNN called Recurrent-Decoder (ERD) that adds fully connected layers before and after an LSTM~\cite{nc1997:Hochreiter} layer and minimizes an Euclidean loss. Their probabilistic aspect uses a Gaussian Mixture Model (GMM) with five mixture components and minimizes the GMM negative log-likelihood. For both aspects, they predict a single future human pose at a time. To predict more, they recurrently feed the single predicted pose back to the input. One drawback of this approach is error drifting, where the prediction error of the current pose will propagate into the next pose. In contrast, we predict multiple human poses at once thus avoiding error drifting. In addition, we do not impose any statistical model constrains like GMM over the motions.

In~\cite{cvpr2016:Ashesh}, the authors develop a general framework that converts a structure graph to an RNN, called a Structure-RNN (S-RNN). They test their framework on different problem sets including human motion prediction, and show that it outperforms the current state of the art. However, they need to design the structure graph manually and task-specifically. In~\cite{cvpr2017:julieta}, the authors examine recent deep RNN methods for human motion prediction, and show that they achieve start-of-the-art results with a simpler model by proposing three simple changes to RNN. In~\cite{corr2017:Judith}, the authors use an encoder-decoder network based on a feed-forward network, and compare the results of three different such architectures: symmetric, time-scale, and hierarchical.

However, the main issues of deterministic prediction of human motion are two-fold. The future is not deterministic, so the same previous poses could lead to multiple possible poses. And, using an $L_{2}$ norm can cause the model to average between two possible futures~\cite{corr2015:Mathieu}, resulting in blurred motions.

\subsection {Non-human motion prediction}

The prediction of multiple possible futures using RNNs has precedents. In \cite{corr2013:Graves}, the authors use an LSTM\cite{nc1997:Hochreiter} to generate text and handwriting from an input sequence. They generate one item at a time, by sampling the resultant probability. Then, they append the predicted item to the input sequence and remove its first item, and iterate. This creates a sequence of desired length, but the method may eventually create an input to the LSTM that does not contain any of the original input. In contrast, we pursue a method that trains our model to generate the entire desired output sequence of poses all at once.

The prediction of single or multiple possible futures using GANs also has precedents. In\cite{corr2015:Mathieu}, the authors trained a convolution model for both the generator and the discriminator in order to predict future frames. They improved the predictions by adding an image gradient difference loss to the adversary loss.  However, they again only predict a single possible future and, due to the use of the convolution network, the model can only predict a fixed length output. In contrast, we support variable length input and variable length output, and can also generate multiple possible futures from the same input.

In \cite{corr2017:Chen}, the authors predict or imagine multiple frames from a single image. They generate affine transformations between each frame, and apply them to the original input image to produce their prediction. Although this can imagine multiple futures from the same input image, a single image is not sufficient to capture the temporal dynamics of a scene. Furthermore, it makes the oversimplified assumption that the change between the images can be captured by using an affine transform.

\subsection {Human motion quality assessment}

Compared to the amount of work on motion editing and synthesis in computer graphics, the research on automatic motion quality evaluation has received little attention~\cite{tog:Ren2005,tog:Harrison2004,tog:Hodgins1998,tog:Reitsma2003}. The existing techniques were typically designed for special types of motions, or for motions obtained from software used to edit character motion~\cite{cmm:Wang2014}. A novel aspect of our motion quality assessment model is that it is trained simultaneously together with the motion prediction model.

\section{Our model for human motion prediction}

In the human motion prediction problem, the system takes a sequence of human poses as input and predicts valid future poses. Let $x=\{x_{1},x_{2},...,x_{m}\}$ be the sequence input poses and $z=\{z_{1},z_{2},...,z_{n}\}$ be the sequence of predicted poses, where each $z_{i}$ and $x_{j}$ corresponds to a single pose, represented by joint locations or joint angles. The goal is to learn the probability of the future sequence conditioned on the input sequence, $P(z|x)$.

\begin{figure*}[ht]
\centering
\includegraphics[width=\textwidth]{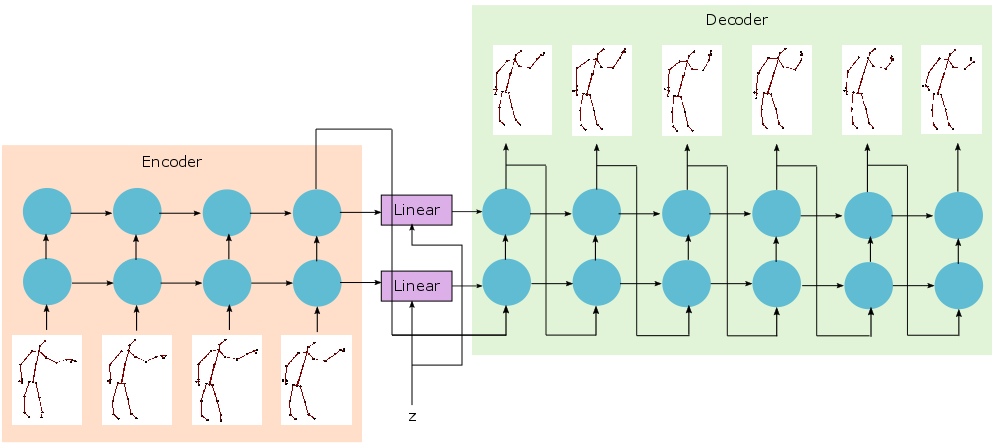}
\caption{Sequence-to-sequence generator network. It predicts multiple future sequences from the same input by feeding in different $z$ values.}
\label{figure:seq-to-seq-gen}
\end{figure*}

As shown in Figure~\ref{figure:seq-to-seq-gen}, our prediction model is a modified version of the sequence-to-sequence~\cite{nips2014:Ilya} network. It takes as input a sequence of human poses, plus a $z$ vector drawn from a uniform or Gaussian distribution $z \sim p_{z}$. The drawn $z$ value is then mapped to the same space as the output states of the encoder. We then simply add the mapped value of $z$ to the encoder states, and use the result as the initial state of the decoder. As shown in the figure, we map $z$ to each layer of the encoder, and then feed the last output of the encoder to the first input of the decoder. We use GRU~\cite{corr2014:Chung} for our sequence-to-sequence network, we also tried LSTM~\cite{nc1997:Hochreiter} and did not notice any quality difference.

Let $G$ be the network shown in figure ~\ref{figure:seq-to-seq-gen}. We have $y = G(x,z;\theta_{g})$, where $\theta_{g}$ are the network parameters that we need to learn. Each drawn value of $z$ will sample different valid future poses from the given input $x$. Most recent human pose predictions using deep RNNs~\cite{iccv2015:Katerina, cvpr2017:julieta} treat human motion prediction as a regression problem. However, solving human motion prediction using regression suffers from two deficiencies. First, it learns one outcome at a time, and as the predicted sequence length increases, this outcome becomes less probable. Second, using the usual $\ell_{1}$ or $\ell_{2}$ loss creates artifacts~\cite{corr2015:Mathieu}, such as predicting the average of two possible outcomes.

We believe GAN is a better way to learn multiple possible future poses from the same input poses. In particular, we use a modified version of the improved WGAN-GP~\cite{corr2017:Ishaan} tailored for sequence prediction, in order to train $y = G(x,z;\theta_{g})$.

\subsection{Generative adversarial networks}

Generative adversarial networks(GAN) was introduced by~\cite{nips2014:Goodfellow}. It is a unsupervised learning technique inspired by the minimax theorem~\cite{pjm1958:Sion}, in which the generator network and the discriminator network try to outdo each other. The training itself alternates between both networks. In the original paper, the generator learns to generate images close to real images, and the discriminator learns to distinguish between the generated image and the real image from the dataset. In a steady state, the discriminator should predict if an image from the generator network is generated or not with $50\%$ probability.

However, the original GAN algorithm is not stable and is difficult to train, because of its use of Jensen-Shannon (JS) divergence as its loss function. JS can result in zero or infinity due to its ratio between two probabilities that might not overlap initially, and can lead to vanishing gradients in the discriminator network. WGAN~\cite{corr2017:Arjovsky} replaces the JS distance with the Earth Mover Distance (EMD), which is defined and continuous almost everywhere. And according to the author, this mitigates the need to carefully maintain a balance between training the discriminator versus the generator. The discriminator in WGAN does not output a probability, and it does not discriminate between synthetic input and real input, which is why the author renamed the discriminator network to critic network.

Nevertheless, WGAN does not address all the concerns, since the critic still must maintain a Lipschitz constraint. In order to do so, the author clips the weights of the network, which adversely affects the quality of the generator. To address this, the WGAN-GP algorithm~\cite{corr2017:Ishaan} replaces weight clipping in the generator with an added penalty to the loss in the critic, which is based on the computed norm of the gradient with respect to the critic input. We were able to verify these improvements in our domain of human poses through extensive experimentation.

Therefore, we use an adversarial training scheme for three reasons. First, it allows the generation of multiple futures from a single past. Second, it allows the generator to be trained without explicitly using the ground truth of real futures. Third, it implicitly learns the cost function for the prediction based on the data. 

\subsection{Human pose prediction GAN (HP-GAN)}
\begin{figure}[ht]
\centering
\includegraphics[width=\linewidth]{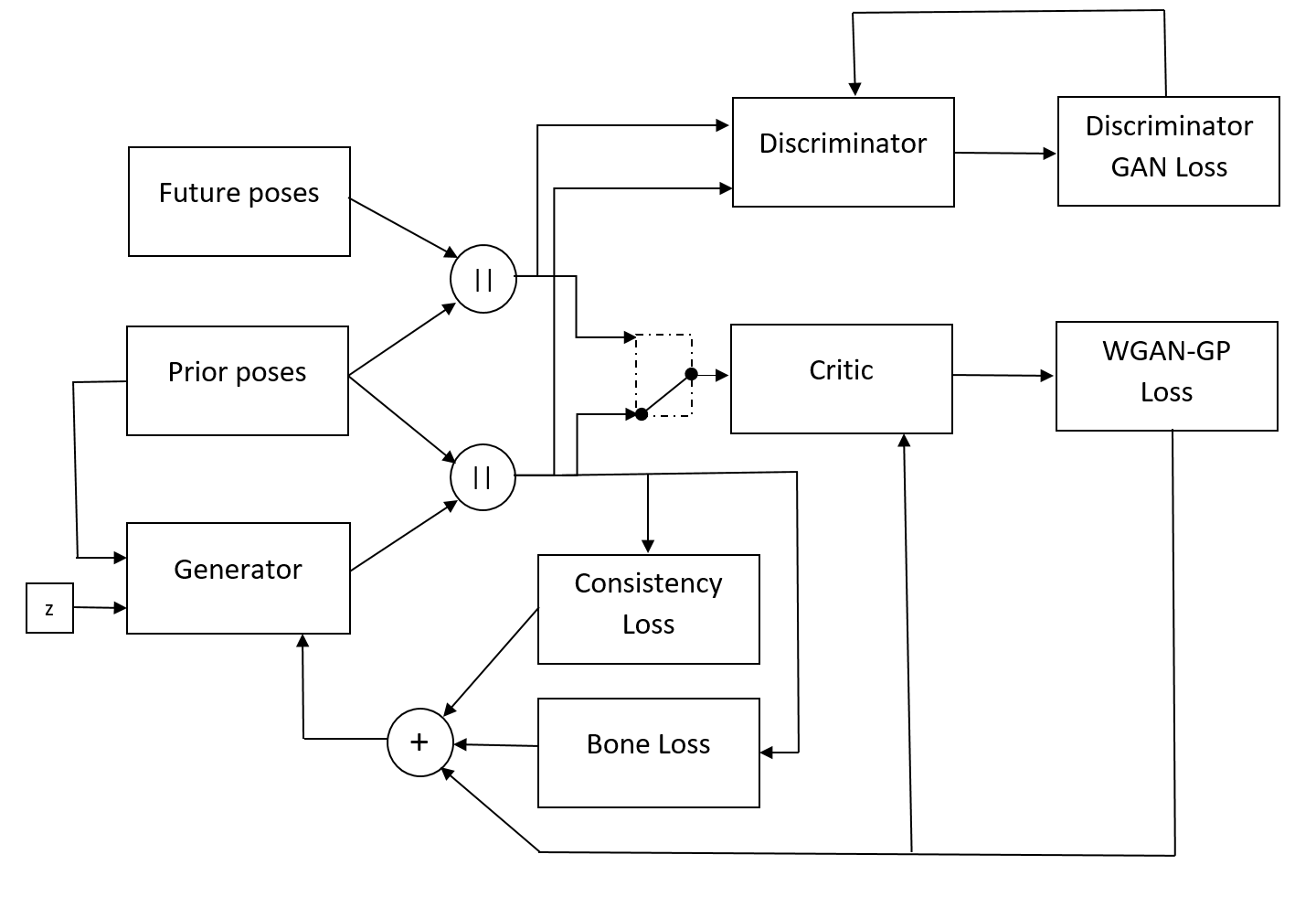}
\caption{HP-GAN: Human Prediction GAN. The above diagram shows the critic switching back and forth between generator and critic training. The output of the generator and the prior poses are concatenated for the critic and discriminator.  Similarly, ground truth is the concatenation of the prior and future poses. The only job of the discriminator is to learn to distinguish between real and fake sequences of human motion. To update the generator, we use WGAN-GP loss, in addition to consistency loss and bone loss.}
\label{figure:seq-to-seq-wgan}
\end{figure}

Figure~\ref{figure:seq-to-seq-wgan} shows the high level diagram of our proposed GAN network, called HP-GAN, for Human Pose prediction. HP-GAN combines features from WGAN-GP~\cite{corr2017:Ishaan}, from GAN~\cite{nips2014:Goodfellow}, and from sequence-specific optimization, in order to generate a realistic human motion sequence, and at the same time to quantify the quality of the generated sequence. 

The "Generator" block shown in Figure~\ref{figure:seq-to-seq-wgan} is the sequence-to-sequence network defined in Figure~\ref{figure:seq-to-seq-gen}. As described earlier, it takes as input previous poses and a $z$ vector, and produces a sequence of human poses. The $z$ vector is a 128-dimensional float vector drawn from a uniform or Gaussian probability distribution. "Future poses" are the ground truth future poses from the dataset, and "Prior poses" are their corresponding previous poses.

Prior poses and future poses are concatenated together to form a real pose sequence. Similarly, prior poses and generated poses are concatenated together to form a fake pose sequence. Both real and fake sequences are used for the critic WGAN-GP loss and for the discriminator GAN loss. WGAN-GP loss is used to improve both the critic network and the generator network by alternating between critic loss and generator loss, whereas the discriminator GAN loss is used only to train the discriminator network. The main purpose of the discriminator network is to measure the quality of the generated human poses; it is not involved in training the generator.

The critic network is a three-layer fully connected feedforward network that outputs a single value. This value is unbounded and is used for the WGAN-GP loss. This WGAN-GP loss is the same loss as in~\cite{corr2017:Ishaan}. To train the generator to produce a realistic human pose, we add two additional losses to the WGAN-GP loss. The first one is the consistency, or pose gradient loss, which focuses on smoothing the sequence of predicted poses. The second loss is the bone loss, which focuses on reducing the changes to the bone lengths between the predicted skeleton and the ground truth.  The details of each of the losses are described below.

The discriminator network is also a three-layer fully connected feedforward network that outputs a single value. However, the output is a probability value between 0 to 1, where 1 means that the input sequence is real and 0 means that the input sequence is fake. The primarily goal of the discriminator network is to measure the quality of the predicted future sequence from the generator. The probability resulting from the discriminator is not used to improve the generator, in order to keep training the generator with WGAN-GP loss, for stability and the avoidance of model collapse. 


\subsubsection{Critic loss}

Our critic loss is similar to WGAN-GP critic loss, with an added L2 regularization. It is defined as follows:

\begin{equation}
L_{c}=L_{wgan}+\lambda L_{gp}+\alpha L_{2}
\label{eq:critic_total_loss}
\end{equation}

$L_{wgan}$ is the original WGAN critic loss, which is defined as:

\begin{equation}
L(x,y,z)_{wgan}=D(x||G(x,z))-D(x||y)
\label{eq:critic_loss}
\end{equation}
where $||$ indicates concatenation, $x$ is the input sequence, $y$ is the future sequence, and $z$ is a random vector drawn from a uniform distribution. 

$L_{gp}$ is the gradient penalty loss defined as:

\begin{equation}
L(x,y,z)_{gp}=(\Vert\nabla_{\hat{x}}D(\hat{x})\Vert_{2} - 1)^{2}
\label{eq:gradient_penalty}
\end{equation}
where $\hat{x}=\epsilon (x||y)+(1-\epsilon)(x||G(x,z))$ and $\epsilon
\sim U[0, 1]$.

$L_{2}$ is the standard $L_{2}$ regularization defined as:

\begin{equation}
L_{2}=\Vert\theta_{d}\Vert_{2}
\label{eq:critic_l2}
\end{equation}

In all of our experiments, we set $\gamma = 10$ and $\alpha = 0.001$.

\subsubsection{Generator loss}

For the generator loss, we use three components. The first is the standard WGAN-GP adversary loss. The second is a consistency loss or pose gradient loss, which measures the delta between two consecutive poses. The third is a ``bone loss'', which measures the delta of bone length between the predicted skeleton and the ground truth. The reason for both the consistency loss and the bone loss is that, in a valid human motion, the change in joint locations between two consecutive frames are small and the bone lengths should remain the same.  

In summary, the generator network loss is as follows.

\begin{equation}
L_{g}=L_{adv}+\alpha L_{pg}+\beta L_{b}
\end{equation}
where $L_{adv}$ is the generator loss in WGAN-GP defined as:

\begin{equation}
L(x,z)_{adv}=-D(x||G(x,z))
\label{eq:g_loss}
\end{equation}
and $L_{pg}$ is the consistency loss, or pose gradient loss, defined as:

\begin{equation}
L(x,z)_{pg}=\Vert\nabla_{t}y\Vert_{p} = \left[\sum_{t}\vert y_{t}-y_{t-1}\vert^{p}\right]^{1/p}
\label{eq:pose_gradient_loss}
\end{equation}
In equation~\ref{eq:pose_gradient_loss}, $\Vert\nabla_{t}y\Vert_{p}$ computes the gradient over time for the predicted sequence, we use $p=2$ in our training. And $L_{b}$ is simply the $L_{2}$ norm of the bone length differences between the predicted pose and the ground truth, that is,

\begin{equation}
L_{b}=\sum_{t}\left[\sum_{i}\vert b^{i}_{t}-b^{i}_{gt}\vert^{2}\right]^{1/2}
\label{eq:bone_loss}
\end{equation}
where $b^{i}_{gt}$ is the ground truth bone length and $b^{i}_{t}$ is the predicted bone length, both at time $t$. We iterate through all bones using index $i$, and sum over all the future skeleton poses using index $t$.

The effect of $L_{pg}$ is critical during training. If $\lambda$ is too large, the effect of $L_{pg}$ loss becomes too high and we obtain less variety in future predictions; in some cases we even obtain identical predictions regardless of the input $z$ value. Conversely, we have found that $L_{pg}$ is not critical if we train on a large dataset, like all the 60 classes of NTURGB-D; but for small subset, like one or two classes, then its value becomes critical in avoiding motion discontinuities at the first predicted human pose. Furthermore, to make sure that $L_{pg}$ loss does not reach zero, we set a minimum value for the loss as follow $L(x,z)_{pg}=max(C,\Vert\nabla_{t}y\Vert_{p})$, where $C$ is a hyper parameter, the reason is to avoid pushing two consecutive human poses to match exactly. 

\subsubsection{Discriminator loss}
We use the discriminator to judge the quality of the predicted skeleton poses and to decide when to stop the training. The discriminator loss is the same loss used in GAN~\cite{nips2014:Goodfellow}, with the exception that the generator does not use this loss in its training. It is defined as:

\begin{equation}
L_{d}=L_{GAN}+\alpha L_{2}
\label{eq:discriminator_total_loss}
\end{equation}
where $L_{GAN}$ is the standard GAN loss defined as:

\begin{equation}
L_{GAN}=log(D(x||y))+log(1-D(x||G(x,z)))
\label{eq:discriminator_loss}
\end{equation}
and $L_{2}$ is the same $L_{2}$ norm used by the critic network.

\subsection{Training}
For the training algorithm, we closely follow most GAN training methods. Inside the training loop, we iterate $k$ times on the critic network, and one time on the generator and discriminator network. We use $k = 10$. We have tried different iteration values, and have tried to dynamically update the iteration count based on the losses of the critic and the generator, but none of those methods made any noticeable improvement.

In order to make the discriminator training procedure more stable, we reduced its learning rate by half, compared to the critic and generator learning rate. For all three networks, we use ADAM~\cite{corr2014:Kingma}, and set the learning rate as $5e-5$ for the critic and generator network, and half of that for the discriminator network.

\section{Experiments}
\begin{figure}[H]
\centering
\includegraphics[width=\linewidth]{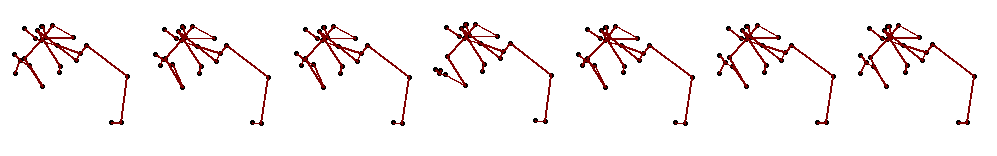}
\caption{Example of bad data from NTURGB-D dataset.}
\label{figure:ntu-bad-data}
\end{figure}

To verify our model capability, we run multiple experiments on two of the largest human motion datasets: a Microsoft Kinect dataset NTURGB-D~\cite{cvpr2016:Shahroudy} and a motion capture (MoCap) dataset Human3.6M~\cite{pami2014:Ionescu,iccv2011:Ionescu}. The human poses in NTURGB-D dataset are based on skeleton data from Kinect, which are not perfect as shown in figure~\ref{figure:ntu-bad-data}, and have problems due to occlusions, or to people carrying objects or interacting with another person. However, even with noisy skeletons our model generalizes well on this dataset.

The NTURGB-D action recognition dataset consists of 56,880 actions, and each action comes with the corresponding RGB video, depth map sequence, 3D skeletal data, and infrared video. We use only the 3D skeleton data. They contain the 3D locations of 25 major body joints at each frame, as defined by the Microsoft Kinect API. NTURGB-D has 60 action classes and 40 different subjects, and each action was recorded by three Kinects from different viewpoints.

Human3.6M contains 3.6 million 3D human poses and their corresponding images, captured by a Vicon MoCap system. Each of these skeletons has 32 joints. The actions were performed by 11 professional actors covering 17 action classes. Using the code from~\cite{cvpr2017:julieta}, we read the Human3.6m skeleton data and converted it from its angle representation to absolute 3D joint positions.

So for both datasets, we trained our model directly on the absolute 3D joint locations. We fed our model a 3D point cloud, and from this training data the model learns the relationships between the joints in order to predict a valid human pose. This is more difficult than training on the angle, which has less degrees of freedom. We train directly on the joint positions in order to use the same pipeline for both NTU-RGB-D and Human3.6m datasets, and to have a more generic model.

\subsection{Pre-processing}

For preprocessing, we normalized the (x,y,z) values of each joint to the range [-1,1] and subtracted the center of gravity. In the NTURGB-D dataset, we normalized each joint to the range of [-1, 1] by using the dimensions of the Kinect frustum at its maximum 5 meter depth. For the Human3.6M dataset, we obtained the range of the raw data by first finding the minimum and maximum values on each dimension, and then computing the minimum and maximum values over all the dimensions.

The Human3.6M dataset has fewer clips than the NTURGB-D dataset, however each clip is much longer. In order to use the same pipeline for both datasets, we split Human3.6M clips into shorter segments and only use every other frame in our training.

\subsection{Quantifying the results}

One of the main problems of GANs is that the loss does not provide any indication of the quality of the generated data. According to the authors of WGAN~\cite{corr2017:Arjovsky} and WGAN-GP~\cite{corr2017:Ishaan}, one of the improvements that WGAN made on the original GAN is that their loss value does in fact provide a quality measure. In our human motion prediction problem, the loss provides some indication of how much the generated sequence looks like a valid human pose. However, we have observed that this is not strictly monotonic; a smaller loss does not always indicate a better quality.

Therefore, to more usefully measure quality, we added a discriminator network, whose sole purpose is to learn the probability that a given sequence is a valid human motion. To find the best model, inside the training loop we generate $N$ predictions and compute the probability, by evaluating the discriminator on each of those predictions. Then, we compute the number of predictions $k$ that has probability more than 50\%. We keep track of the model with maximum $k$ during training, we only start tracking the best model after certain number of epochs, in order to give the discriminator a chance to learn to discriminate between a real and a fake human motion.  

\section{Results}
\begin{figure*}[ht]
\centering
\includegraphics[width=\textwidth]{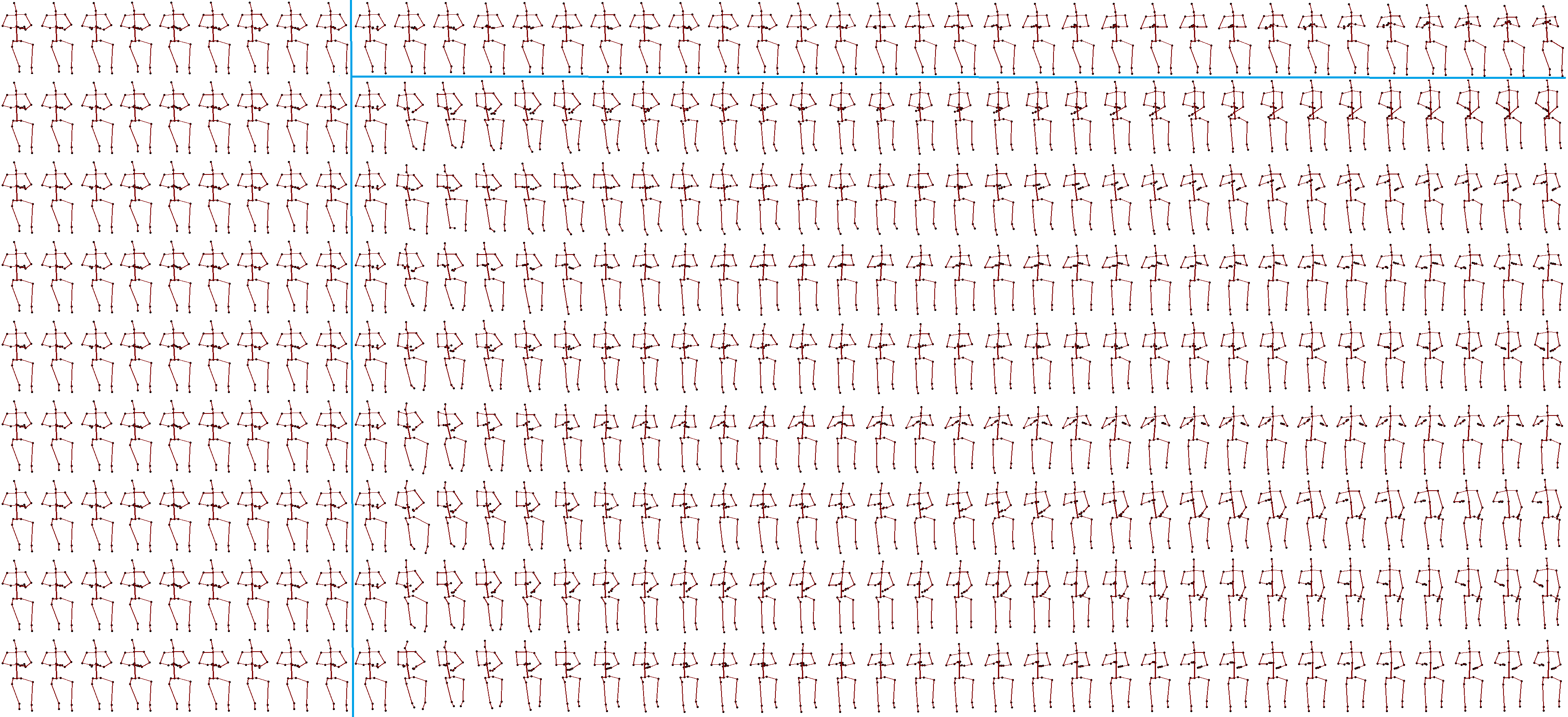}
\caption{Training on all 60 classes from the NTURGB-D dataset, and then predicting 30 poses from 10 input poses. Top row is the ground truth, and each subsequent row corresponds to a random $z$ value drawn from a uniform distribution. Poses to the left of the blue line are the input sequences. Pose to the right of the blue line, starting in the second row, are the predicted future poses for each $z$ value.}
\label{figure:ntu_all_pred_184}
\end{figure*}

Figure~\ref{figure:ntu_all_pred_184} shows the result of training HP-GAN on all 60 classes of the NTURGB-D dataset. The top row is the ground truth, and each subsequent row corresponds to the predicted human poses from a different $z$ value drawn from a uniform distribution. To the left of the blue line are the input sequences. As shown in the figure, each $z$ value generates a separate possible future sequence of human poses. The first few of the predicted poses are very close to the ground truth, which is expected. As we predict more poses, they start to differ more from the ground truth. These results did not use the consistency loss, which is more important for smaller datasets or for training on a subset of the actions. We can see that there is no discontinuity between the last input pose and the first predicted pose. 

\begin{figure}[H]
\centering
\includegraphics[width=\linewidth]{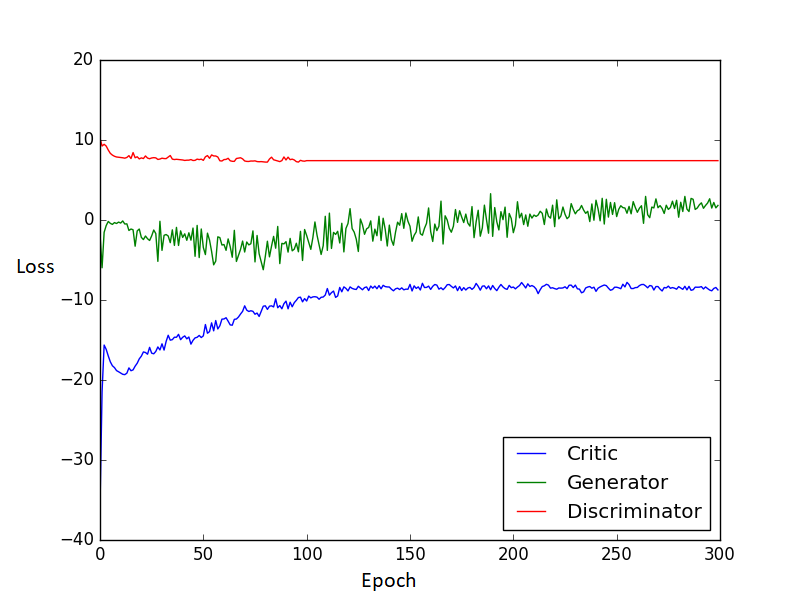}
\caption{Critic versus Generator versus Discriminator loss.}
\label{figure:loss-plot}
\end{figure}

In figure~\ref{figure:loss-plot}, we plot the loss values as a function of the number of epochs, for the critic (blue), generator (green), and discriminator (red) losses. Even though the generator loss continues improving, it is difficult to compute the best parameters for either generator loss or critic loss without a visual inspection of their graphs. The discriminator loss provides a more stable signal.

\begin{figure}[ht]
\centering
\includegraphics[width=\linewidth]{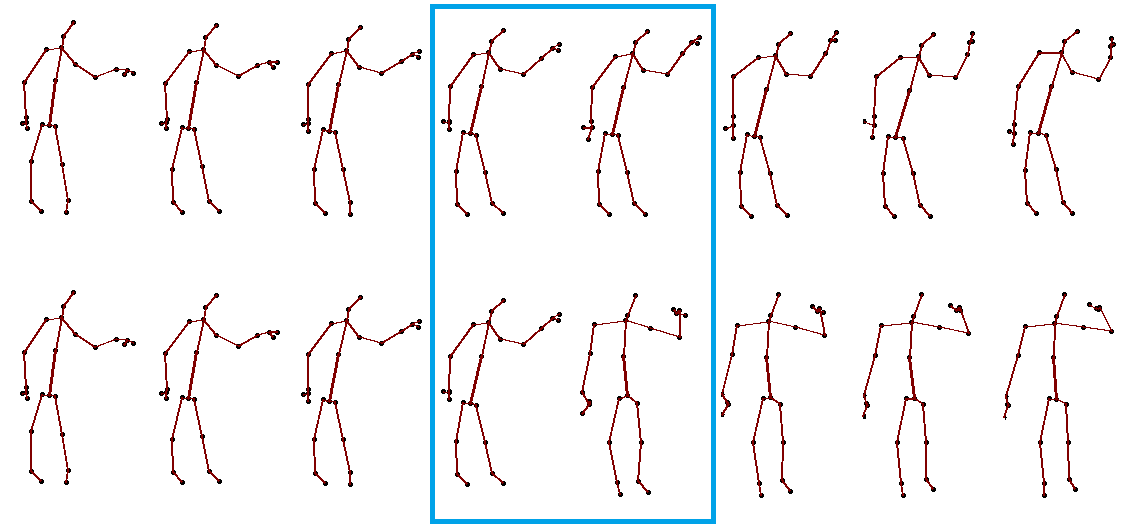}
\caption{Discontinuity between input poses and predicted poses. Top row is ground truth and bottom row is the generated sequence.}
\label{figure:jumpiness}
\end{figure}

Not all generated human poses are smooth, especially when we train on a subset of the actions. Figure~\ref{figure:jumpiness} shows a discontinuity between the last pose of the input and the first pose of the prediction. This can be mitigated by increasing the effect of the consistency loss in the generator loss, or by reducing the capacity of the network, or by feeding the actual last input directly into the first decoder node.

\begin{figure}[H]
\centering
\includegraphics[width=\linewidth]{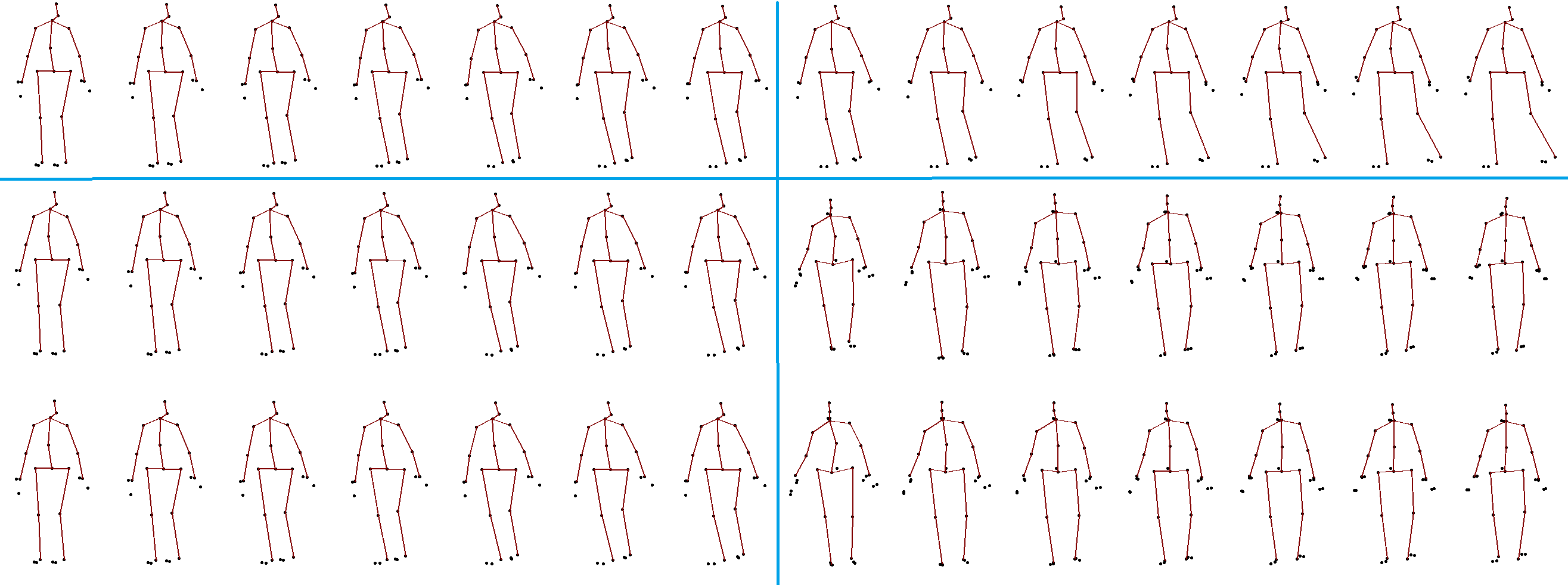}
\caption{Human3.6M example. Top row is the ground truth. Next two rows: to left of line is the input sequence, to right of line are the predicted sequences from two different $z$ values.}
\label{figure:human36m}
\end{figure}

Training on the Human3.6M dataset requires different adjustments from training on NTURGB-D dataset. For example, bone loss is more important for the Human3.6M dataset than for the NTU-RGB-D dataset. Since NTU-RGB-D skeletons are generated from the Kinect, bone lengths between poses for the same subject do not match exactly. Also, Human3.6M actions are slower than NTU-RGB-D actions, so there is little variation between poses predictions even with different $z$ values. To mitigate this, we only use every other frame. Figure~\ref{figure:human36m} shows results on the Human3.6M dataset.

\begin{figure}[H]
\centering
\includegraphics[width=\linewidth]{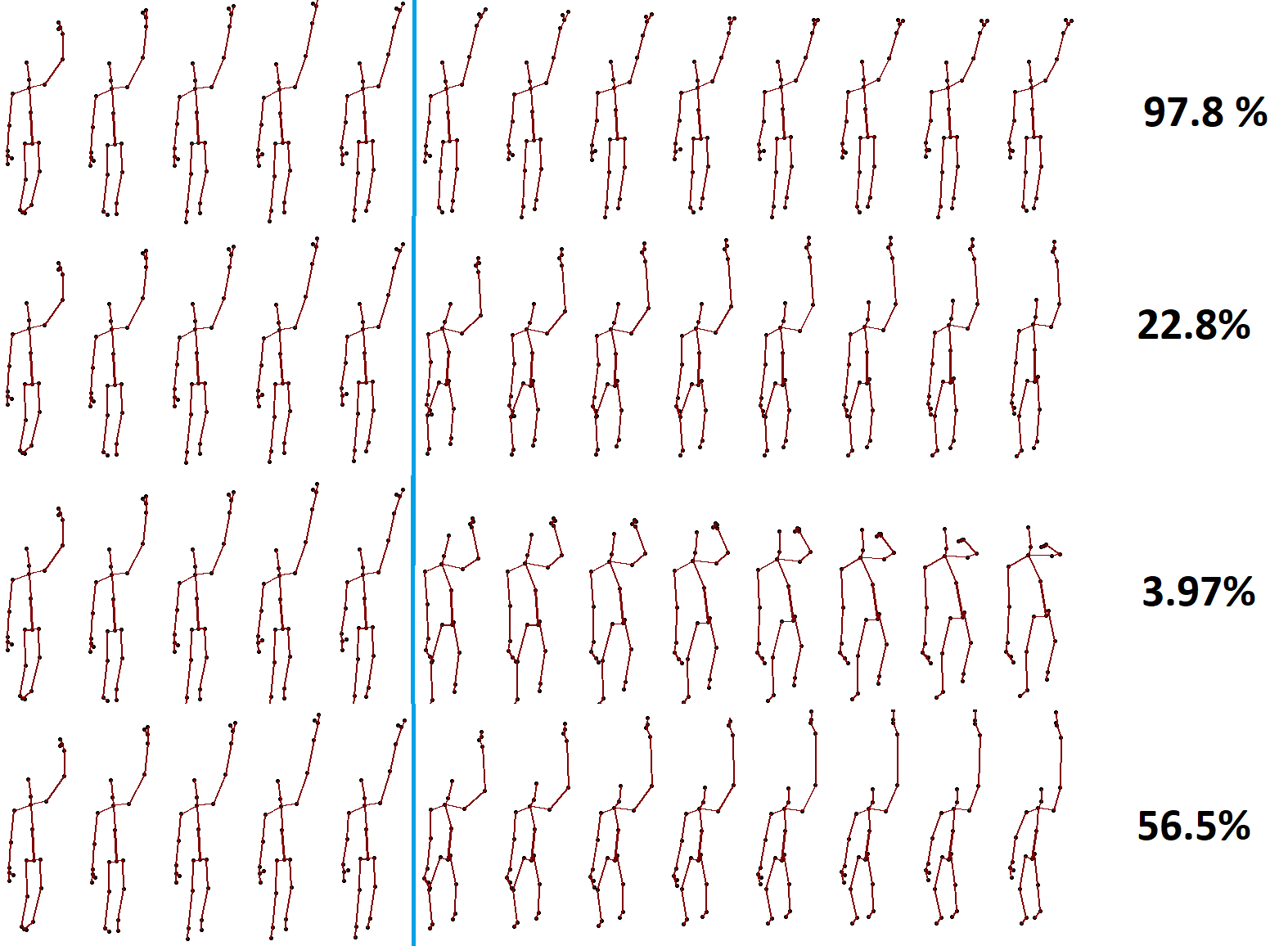}
\caption{Sequences for the "throw" action, labeled with their corresponding probability from the discriminator network. Top row is ground truth.}
\label{figure:probability}
\end{figure}

Figure~\ref{figure:probability} shows predictions of the "throw" action from the NTURGB-D dataset and their corresponding estimated probabilities from the discriminator network. The top row is the ground truth, which shows a 97.8\% probability of being a valid human sequence and each subsequent row correspond to a different $z$ value. The discriminator probability output is not always accurate especially at the start of the training. We tried to use the discriminator in addition to the critic to update the generator network parameters, however, the result was worse and the training was less stable.

\section{Conclusions and future work}

We have developed a novel sequence-to-sequence model for probabilistic human motion prediction, which predicts multiple plausible future human poses from the same input. To our knowledge, we are the first to use deep neural networks for probabilistic motion predictions. To quantify the quality of the non-deterministic predictions, we simultaneously trained a motion-quality-assessment model that learns the probability that a given skeleton sequence is a real human motion. We tested our architecture on two different datasets, one based on the Kinect sensor and the other based on MoCap data. Experiments show that our model performs well on both datasets.

However, using WGAN-GP and all the improvements we added, there still no good way to tell if the training has converged. Even worse, the training can diverge after it already converged if we continue the training loop. Therefore, in future work, we plan to explore better mechanism to measure model convergence and to ensure stability. Another area of research, is understanding the semantic and space of the $z$ vector. If we can compute the reverse mapping from sequences to $z$, we might be able to use $z$ values for action classification or clustering. Furthermore, HP-GAN generates a new dataset that is not in the original source, so an area of exploration is to use HP-GAN for data augmentation in order to increase diversity per action. 

Lastly, we used standard GRU in our sequence-to-sequence network and we did not investigate new RNN architecture that might provide better human pose prediction such as RNN with skip connection~\cite{cvpr2017:julieta} or Part-Aware LSTM~\cite{cvpr2016:Shahroudy} or something else, this is another area we plan to investigate in the future.

\clearpage

{\small
\bibliographystyle{ieee}
\bibliography{references}
}

\end{document}